\documentclass[letterpaper]{article}
\usepackage{helvet}
\usepackage{courier}
\usepackage[latin9]{inputenc}
\usepackage{float}
\usepackage{amsmath}
\usepackage{graphicx}
\usepackage{cite}
\makeatletter

\providecommand{\tabularnewline}{\\}
\floatstyle{ruled}
\newfloat{algorithm}{tbp}{loa}
\providecommand{\algorithmname}{Algorithm}
\floatname{algorithm}{\protect\algorithmname}

\def\year{2017}\relax
\usepackage{aaai17}
\usepackage{times}
\usepackage{url}
\usepackage{algorithm,algpseudocode}
\usepackage{amsfonts}
\nocopyright

\makeatother

\begin{document}

\title{Inverse Classification for Comparison-based Interpretability in Machine
Learning}

\author{Thibault Laugel\textsuperscript{1}, Marie-Jeanne Lesot\textsuperscript{1},
Christophe Marsala\textsuperscript{1}, Xavier Renard\textsuperscript{2},
Marcin Detyniecki\textsuperscript{1,2,3}\\
 \textsuperscript{1}Sorbonne Universit�s, UPMC Univ Paris 06, CNRS,
LIP6 UMR 7606, 4 place Jussieu 75005 Paris, France\\
 \textsuperscript{2}AXA \textendash{} Data Innovation Lab, 48 rue
Carnot 92150 Suresnes, France\\
\textsuperscript{3}Polish Academy of Science, IBS PAN, Warsaw, Poland\\
thibault.laugel@lip6.fr }
\maketitle
\begin{abstract}
In the context of post-hoc interpretability, this paper addresses
the task of explaining the prediction of a classifier, considering
the case where no information is available, neither on the classifier
itself, nor on the processed data (neither the training nor the test
data). It proposes an instance-based approach whose principle consists
in determining the minimal changes needed to alter a prediction: given
a data point whose classification must be explained, the proposed
method consists in identifying a close neighbour classified differently,
where the closeness definition integrates a sparsity constraint. This
principle is implemented using observation generation in the Growing
Spheres algorithm. Experimental results on two datasets illustrate
the relevance of the proposed approach that can be used to gain knowledge
about the classifier.
\end{abstract}

\section{Introduction}

Bringing transparency to machine learning models is nowadays a crucial
task. However, the complexity of today's best-performing models as
well as the subjectivity and lack of consensus over the notion of
interpretability make it difficult to address.

Over the past few years, multiple approaches have been proposed to
bring interpretability to machine learning, relying on intuitions
about what 'interpretable' means and what kind of explanations would
help a user understand a model or its predictions. Existing categorizations
of interpretability \cite{Bibal2016a,Kim2017,Doshi-Velez2017,Cotton2008}
usually distinguish approaches mainly on the characteristics of these
explanations: given a classifier to be interpreted, \textit{in-model}
interpretability relies on modifying its learning process to make
it simpler (see for instance Abdollahi et al. 2016)\nocite{Abdollahi2016}.
Other approaches consist in building a new simpler model to replace
the original classifier \cite{Lakkaraju2017,Angelino2017}. On the
contrary,\textit{ post-hoc} interpretability focuses on building an
explainer system using the results of the classifier to be explained.

In this work, we propose a post-hoc approach that aims at explaining
a single prediction of a model through comparison. In particular,
given a classifier and an observation to be interpreted, we focus
on finding the closest possible observation belonging to a different
class. 

Explaining through particular examples has been shown in cognitive
and teaching sciences to facilitate the learning process of a user
(see e.g. Watson et al. (2008)). This is especially relevant in cases
where the classifer decision to explain is complex and other interpretability
approaches cannot provide meaningful explanations. Another motivation
for our approach lies in the fact that in many applications of machine
learning today, no information about the original classifier or existing
data is made available to the end-user, making model- and data-agnostic
intepretability approaches essential. 

To address these issues we propose \textit{Growing Spheres}, a generative
approach that locally explores the input space of a classifier to
find its decision boundary. It has the specifity of not relying on
any existing data other than the observation to be interpreted to
find the minimal change needed to alter its associated prediction.

The paper is organized as follows: we first present some existing
approaches for post-hoc interpretability and how they relate to the
one proposed in this paper. Then, we describe the proposed comparison-based
approach as well as its formalization and motivations. We then describe
the \textit{Growing Spheres} algorithm. Finally, we illustrate the
method through two real-world applications and analyze how it can
be used to gain information about a classifier.

\section{Post-hoc Interpretability}

Post-hoc interpretability approaches aim at explaining the behavior
of a classifier around particular observations to let the user understand
their associated predictions, generally disregarding what the actual
learning process of the model might be. Post-hoc interpretability
of results has received a lot of interest recently (see for instance
Kim and Doshi-Velez (2017)), especially as black-box models such as
deep neural networks and ensemble models are being more and more used
for classification despite their complexity. This section briefly
reviews the main existing approaches, depending on the hypotheses
that are made about available inputs and on the forms the explanations
take. These two axes of discussion, which obviously overlap, provide
a good framework for the motivations of our approach.

\subsection{Available Inputs}

Let us consider the case of a physician using a diagnostic tool. It
is natural to speculate that (s)he does not have any information about
the machine learning model used to make disease predictions, neither
may (s)he have any idea about what patients were used to train it.
This raises the question of what knowledge (about the machine learning
model and the training or other data) an end-user has, and hence what
inputs a post-hoc explainer should use. 

Several approaches rely specifically on the knowledge of the algorithm
used to make predictions, taking advantage of the classifier structure
to generate explanations \cite{Barbella2009,Hendricks2016}. However,
in other cases, no information about the prediction model is available
(the model might be accessible only through an API or a software for
instance). This highlights the necessity of having model-agnostic
interpretability methods that can explain predictions without making
any hypotheses on the classifier \cite{Baehrens2010,Adler2016,Ribeiro2016}.
These approaches, sometimes called \textit{sensitivity analyzes, }generally
try to analyze how the classifier locally reacts to small perturbations.
For instance, Baehrens et al. (2010) approximate the classifier with
Parzen windows to calculate the local gradient of the model and understand
what features locally impact the class change.

\subsection{Forms of Explanations}

Beyond the differences regarding their inputs, the variety of existing
methods also comes from the lack of consensus regarding the definition,
and a fortiori the formalization, of the very notion of interpretability.
Depending on the task performed by the classifier and the needs of
the end-user, explaining a result can take multiple forms. Interpretability
approaches hence rely on the following assumptions to design explanations: 
\begin{enumerate}
\item The explanations should be an accurate representation of what the
classifier is doing.
\item The explanations should be understandably read by the user. 
\end{enumerate}
Feature importances \cite{Baehrens2010,Ribeiro2016}, binary rules
\cite{Turner2015} or visualizations \cite{Krause2016b} for instance
give different insights about predictions without any knowledge on
the classifier. The LIME approach \cite{Ribeiro2016} linearily approximates
the local decision boundary of a classifier and calculates the linear
coefficients of this approximation to give local feature importances,
while Hendricks et al. (2016) identify class-discriminative properties
that justify predictions and generate sentences to explain image classification. 

In this paper, we consider the case of instance-based approaches,
which bring interpretability by comparing an observation to relevant
neighbors \cite{Mannino,Strumbelj2009,Martens2014a,Kabra2015}. These
approaches use other observations (from the train set, from the test
set or generated ones) as explanations to bring transparency to a
prediction of a black-box classifier. 

One of the motivations for instance-based approaches lies in the fact
that in some cases the two objectives $1$ and $2$ mentioned above
are contradictory and cannot be both reached in a satisfying way.
In these complex situations, finding examples is an easier and more
accurate way to describe the classifier behavior than trying to force
a specific inappropriate explanation representation, which would result
in incomplete, useless or misleading explanations for the user. 

As an illustration, Baehrens et al. (2010) discuss how their approach
based on Parzen windows does not succeed well in providing explanations
for individual predictions that are at the boundaries of the training
data, giving explanation vectors (gradients) actually pointing in
the (wrong) opposite direction from the decision boundary. Comparison
with observations from the other class would probably make more sense
in such a case and give more useful insights. 

Existing instance-based approaches moreover often rely on having some
prior knowledge, be it about the machine learning model, the train
dataset, or other labelled instances. For instance, Kabra et al. (2015)
try to identify which train observations have the highest direct influence
over a single prediction.

\section{Comparison-based Interpretability}

In this section, we motivate the proposed approach in the light of
the two axes of discussion presented in the previous section.

\subsection*{Explaining by Comparing}

Disposing of knowledge on the classifier or data is an asset existing
methods can use to create the explanations they desire. However, the
democratization of machine learning implies that in a lot of nowadays
cases, the end-user of an explainer system does not have access to
any of this knowledge, making such approaches unrealistic. In this
context, the need for a comparison-based interpretability tool that
does not rely on any prior knowledge, including any existing data,
constitutes one of the main motivations for our work.

Due to its highly subjective nature, interpretability in machine learning
sometimes looks up to cognitive sciences for a justification for building
explanations (when they do not, they rely on intuitive ideas about
what interpretable means). Although not mentioned by the previously
cited instance-based approaches, it must be underlined that learning
through examples also possesses a strong justification in cognitive
and teaching sciences \cite{Deyck1994,Watson2008,Mvududu2011,VanGog2011}.
For instance, Watson et al. (2008) show through experiences that generated
examples help students 'see' abstract concepts that they had trouble
understanding with more formal explanations.

Driven by this cognitive justification and the need to have a tool
that can be used when the available information is scarce, we propose
an instance-based approach relying on comparison between the observation
to be interpreted and relevant neighbors. 

\subsection{Principle of the Proposed Approach}

In order to interprete a prediction through comparison, we propose
to focus on finding an observation belonging to the other class and
answer the question: 'Considering an observation and a classifier,
what is the minimal change we need to apply in order to change the
prediction of this observation?'. This problem is similar to inverse
classification \cite{Mannino}, but we apply it to interpretability.

Explaining how to change a prediction can help the user understand
what the model considers as locally important. However, compared to
feature importances which are often built to have some kind of statistical
robustness, this approach does not claim to bring any causal knowledge.
On the contrary, it gives local insights disregarding the global behavior
of the model and thus differs from other interpretability approaches.
For instance, Ribeiro et al. (2016) evaluate their method LIME by
looking at how faithful to the global model the local explainer is.
However, despite not providing any causal information, the proposed
approach provides the exact values needed to change the prediction
class, which is also very helpful to the user. 

Furthermore, it is important to note that our primary goal here is
to give insights about the classifier, not the reality it is approximating.
This approach thus aims at understanding a prediction regardless of
whether the classifier is right or wrong, or whether or not the observations
generated as explanations are absurd. This characteristic is shared
with adversarial machine learning \cite{Tygar2011,Szegedy2014,Goodfellow2015},
which relates to our approach since it aims at 'fooling' a classifier
by generating close variations of original data in order to change
their predictions. These adversarial examples rely on exploiting weaknesses
of classifiers such as their sensitivity to unknown data, and are
usually generated using some knowledge of the classifier (such as
its loss function). The approach we propose also relies on generating
observations that might not be realistic but without any knowledge
about the classifier whatsoever and for the purpose of interpretability.

\subsection{Finding the Closest Ennemy}

For simplification purposes, we propose a formalization of the proposed
approach for binary classification. However, it can be applied to
multiclass classification.

Let us consider a problem where a classifier $f$ maps some input
space $\mathcal{X}$ of dimension $d$ to an output space $\mathcal{Y}=\{-1,1\}$,
and suppose that no information is available about this classifier.
Suppose all features are scaled to the same range. Let $x=(x_{i})_{i}\in\mathcal{X}$
be the observation to be interpreted and $f(x)\in\mathcal{Y}$ its
associated prediction. The goal of the proposed instance-based approach
is to explain $x$ through an other observation $e\in\mathcal{\mathcal{X}}$.
The final form of explanation is the difference vector $e-x$.

In particular, we focus on finding an observation $e$ belonging to
a different class than $x$, i.e. such that $f(e)\neq f(x)$. For
simplification purposes, we call \textit{ally} an observation belonging
to the same class as $x$ by the classifier, and \textit{ennemy} if
it is classified to the other class.

Recalling objective $1$ mentioned earlier, the final explanation
$e-x$ we are looking for should be an accurate representation of
what the classifier is doing. This is why we decide to transform this
problem into a minimization problem by defining the function $c:\mathcal{X}\times\mathcal{X}\rightarrow\mathbb{R}\text{\textsuperscript{+}}$
such that $c(x,e)$ is the cost of moving from observation $x$ to
ennemy $e$.

Using this notation, we focus on solving the following minimization
problem:

\begin{equation}
e^{*}=\underset{e\in\mathcal{X}}{{\arg\min}}{\{c(x,e)\quad|\quad f(e)\neq f(x)\}}
\end{equation}

The difficulty of defining the cost function $c$ comes from the fact
that despite the classifier being designed to learn and optimize some
specific loss function, the considered black-box hypothesis compells
us to choose a different metric.

Thus, we define $c$ as:

\begin{equation}
c(x,e)=||x-e||_{2}+\gamma||x-e||_{0}
\end{equation}
with $||e-x||_{0}=\sum_{i\leq d}1_{x_{i}\neq e_{i}}$,$\gamma\in\mathbb{R}^{+}$
the weight associated to the vector sparsity and $||.||_{2}$ the
Euclidean norm.

Looking up to Strumbelj et al. (2009), we choose to use the $l_{2}$
norm of the vector $e-x$ as a component of the cost function to measure
the proximity between $e$ and $x$. However, recalling objective
$2$, we need to make sure that this cost function guarantees a final
explanation that can be easily read by the user. In this regard, we
consider that human users intuitively find explanations of small dimension
to be simpler. Hence, we decide to integrate vector sparsity, measured
by the $l_{0}$ norm, as another component of the cost function $c$
and combine it with the $l_{2}$ norm as a weighted average.

Due to the cost function $c$ being discontinuous and the hypotheses
made (black-box classifier and no existing data) solving problem (1)
is difficult. Hence, we choose to solve sequentially the two components
of the cost function using \emph{Growing Spheres}, a two-step heuristic
approach that approximates the solution of this problem. 

\section{Growing Spheres}

In order to solve the problem defined in Equation (1), the proposed
approach \emph{Growing Spheres} uses instance generation without relying
on existing data. Thus, considering an observation to interprete,
we ignore in which direction the closest classifier boundary might
be. In this context, a greedy approach to find the closest ennemy
is to explore the input space $\mathcal{X}$ by generating instances
in all possible direction until the decision boundary of the classifier
is crossed, thus minimizing the $l_{2}$-component of our metric.
This step is detailed in the next part, Generation. 

Then, in order to make the difference vector of the closest ennemy
sparse, we simplify it by reducing the number of features used when
moving from $x$ to $e$ (thus minimizing the $l_{0}$ component of
the cost function and generating the final solution $e^{*}$), as
explained in the Feature Selection part.

\begin{figure}[t]
\begin{centering}
\includegraphics[scale=0.3]{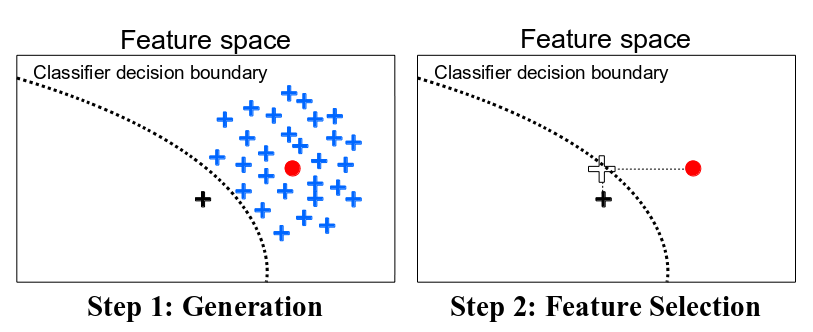} 
\par\end{centering}
\caption{Illustration of Growing Spheres: The red circle represents the observation
to interprete, the plus signs observations generated by Growing Spheres
(blue for allies, black for ennemies). The white plus is the final
ennemy $e^{*}$ used to generate explanations.}
\label{growingspheres}
\end{figure}

An illustration of the two steps of \emph{Growing Spheres} is drawn
in Figure \ref{growingspheres}.

\subsection{Generation}

The generation step of \emph{Growing Spheres} is detailed in Algorithm
\ref{algo_growingspheres}. Its main idea is to generate observations
in the feature space in $l_{2}$-spherical layers around $x$ until
an ennemy is found. For two positive numbers $a_{0}$ and $a_{1}$,
we define a $(a_{0},a_{1})$-spherical layer $SL$ around $x$ as:

\[
SL(x,a_{0},a_{1})=\{z\in\mathcal{X}\;:\;a_{0}\leq||x-z||_{2}\leq a_{1}\}
\]

To generate uniformly over these subspaces, we use the YPHL algorithm
\cite{Harman2010} which generates observations uniformly distributed
over the surface of the unit sphere. We then draw $\mathcal{U}_{[a_{0,}a_{1}]}$-distributed
values and use them to rescale the distances between the generated
observations and $x$. As a result, we obtain observations that are
uniformly distributed over $SL(x,a_{0},a_{1})$.

The first step of the algorithm consists in generating uniformly $n$
observations in the $l_{2}$-ball of radius $\eta$ and center $x$,
which corresponds to $SL(x,0,\eta)$ (line 1 of Algorithm \ref{algo_growingspheres}),
with $n$ and $\eta$ hyperparameters of the algorithm.

In case this initial generation step already contains ennemies, we
need to make sure that the algorithm did not miss the closest decision
boundary. This is done by updating the value of the initial radius:
$\eta\leftarrow\eta/2$ and repeating the initial step until no ennemy
is found in the intial ball $SL(x,0,\eta)$ (lines 2 to 5).

However, if no ennemy is found in $SL(x,0,\eta)$, we update $a_{0}$
and $a_{1}$ using $\eta$, generate over $SL(x,a_{0},a_{1})$ and
repeat this process until the first ennemy has been found (as detailed
in lines 6 to 11).

In the end, Algorithm \ref{algo_growingspheres} returns the $l_{2}$-closest
generated ennemy $e$ from the observation to be interpreted $x$
(as represented by the black plus in Figure \ref{growingspheres}).

\begin{algorithm}[t]
\begin{algorithmic}[1] 
\Require{$f: \mathcal{X} \rightarrow \{-1; 1\}$ a binary classifier}
\Require{$x \in \mathcal{X}$ an observation to be interpreted}
\Require{Hyperparameters: $\eta, n$}
\Ensure{Ennemy $e$}

\State{Generate $(z_i)_{i\leq n}$ uniformly in $SL(x, 0, \eta)$}
\While{$\exists \; e \in (z_i)_{i\leq n} \; | \; f(e) \neq f(x)$}
\State{$\eta = \eta / 2$}
\State{Update $(z_i)_{i\leq n}$ by generating uniformly in $SL(x, (0, \eta))$}
\EndWhile

\State{Set $a_0 = \eta$, $a_1=2\eta $}
\While{$\not\exists \; e \in (z_i)_{i\leq n} \; | \; f(e) \neq f(x)$}
\State{$a_0 = a1$}
\State{$a_1 = a1 + \eta$}
\State{Generate $(z_i)_{i\leq n}$ uniformly in $SL(x, a_0, a_1)$}
\EndWhile
\State{\textbf{Return} $e$, the $l_2$-closest generated ennemy from $x$ }
\end{algorithmic}

\caption{Growing spheres generation}
\label{algo_growingspheres}
\end{algorithm}

Once this is done, we focus on making the associated explanation as
easy to understand as possible through feature selection.

\subsection{Feature Selection}

Let $e$ be the closest ennemy found by Algorithm \ref{algo_growingspheres}.
Our second objective is to minimize the $l_{0}$ component of the
cost function $c(x,e)$ defined in Equation (2). This means that we
are looking to maximize the sparsity of vector $e-x$ with respect
to $f(e)\neq f(x)$. To do this, we consider again a naive heuristic
based on the idea that the smallest coordinates of $e-x$ might be
less relevant locally regarding the classifier decision boundary and
should thus be the first ones to be ignored.

The feature selection algorithm we use is detailed in Algorithm \ref{algo featsel}.

\begin{algorithm}[t]
\begin{algorithmic}[2] 
\Require{$f: \mathcal{X} \rightarrow \{-1; 1\}$ a binary classifier}
\Require{$x \in \mathcal{X}$ the observation to be interpreted}
\Require{$e \in \mathcal{X} \; | \; f(e) \neq f(x)$ the solution of Algorithm 1}
\Ensure{Ennemy $e^*$}
\State{Set $e' = e$}
\While{$f(e') \neq f(x)$}
\State{$e^* = e'$}
\State{$i = \underset{j \in [1:d], \; e'_j \neq x_j}{\arg\min}|e'_j-x_j|$}
\State{Update $e'_i = x_i$}
\EndWhile

\State{\textbf{Return} $e^*$}
\end{algorithmic}

\caption{Feature Selection}
\label{algo featsel}
\end{algorithm}

The final explanation provided to interprete the observation $x$
and its associated prediction is the vector $x-e^{*}$, with $e^{*}$
the final ennemy identified by the algorithms (represented by the
white plus in Figure \ref{growingspheres}).

\section{Experiments}

The aforementioned difficulties of working with interpretability make
it often impossible to evaluate approaches and compare them one to
another. 

Some of the existing approaches \cite{Baehrens2010,Ribeiro2016,Doshi-Velez2017}
rely on surveys for evaluation, asking users questions to measure
the extent to which they help the user in performing his final task,
in order to assess some kind of explanation quality. However, creating
reproducible research in machine learning requires to define mathematical
proxies for explanation quality.

In this context, we present illustrative examples of the proposed
approach applied to news and image classification. In particular,
we analyze how the explanations given by \textit{Growing Spheres}
can help a user gain knowledge about a problem or identify weaknesses
of a classifier. Additionally, we check that the explanations can
be easily read by a user by measuring the sparsity of the explanations
found. 

\subsection{Application for News Popularity Prediction}

We apply our method to explain the predictions of a random forest
algorithm over the news popularity dataset \cite{Fernandes2015}.
Given 58 numerical features created from 39644 online news articles
from website Mashable, the task is to predict wether said articles
have been shared more than 1400 times or not. Features for instance
encode information about the format and content of the articles, such
as the number of words in the title, or a measure of the content subjectivity
or the popularity of the keywords used. We split the dataset and train
a random forest classifier (\textbf{RF}) on 70\% of the data. We use
a grid search to look for the best hyperparameters of \textbf{RF}
(number of trees) and test it on the rest of the data (0.70 final
AUC score). We use $\gamma=1$ to define the cost function $c$ and
set the hyperparameters of Algorithm 1 to $\eta=0.001$ and $n=10000$.

\subsubsection{Illustrative Example}

We apply \textit{Growing Spheres} to two random observations from
the test set (one from each class). For instance, let us consider
the case of an article entitled 'The White House is Looking for a
Few Good Coders' (Article 1). This article is predicted to be not
popular by \textbf{RF}. 

The explanation vector given by \textit{Growing Spheres} for this
prediction has 2 non-null coordinates that can be found in Table \ref{exemple_news_1}:
among the articles referenced in Article 1, the least popular of them
would need to have 2016 more shares in order to change the prediction
of the classifier. Additionally, the keywords used in Article 1 are
each associated to several articles using them. For each keyword,
the most popular of these articles would need to have 913 more shares
in order to change the prediction. In other words, Article 1 would
be predicted to be popular by \textbf{RF} if the references and the
keywords it uses were more popular themselves.

On the opposite, as presented in Table \ref{exemple_news_2}, these
same features would need to be reduced for Article 2, entitled ''Intern'
Magazine Expands Dialogue on Unpaid Work Experience' and predicted
to be popular, to change class. Additionally, the feature 'text subjectivity
score' (score between 0 and 1) would need to be reduced by 0.03, indicating
that a slightly more objective point of view from the author would
lead to have Article 2 predicted as being not popular.

\begin{table}
\begin{centering}
\begin{tabular}{|c|c|}
\hline 
Feature & Move\tabularnewline
\hline 
\hline 
Min. shares of referenced articles in Mashable & +2016\tabularnewline
\hline 
Avg. keyword (max. shares) & +913\tabularnewline
\hline 
\end{tabular}
\par\end{centering}
\caption{Output examples of \textit{Growing Spheres} for Article 1, predicted
to be not popular by \textbf{RF}}
\label{exemple_news_1}
\end{table}

\begin{table}
\begin{centering}
\begin{tabular}{|c|c|}
\hline 
Feature & Move\tabularnewline
\hline 
\hline 
Avg. keyword (max. shares) & -911\tabularnewline
\hline 
Min. shares of referenced articles in Mashable & -3557\tabularnewline
\hline 
Text subjectivity & -0.03\tabularnewline
\hline 
\end{tabular}
\par\end{centering}
\caption{Output examples of \textit{Growing Spheres} for Article 2, predicted
to be popular by \textbf{RF}}
\label{exemple_news_2}
\end{table}

\subsubsection{Sparsity Evaluation}

\begin{figure}[t]
\begin{centering}
\includegraphics[scale=0.4]{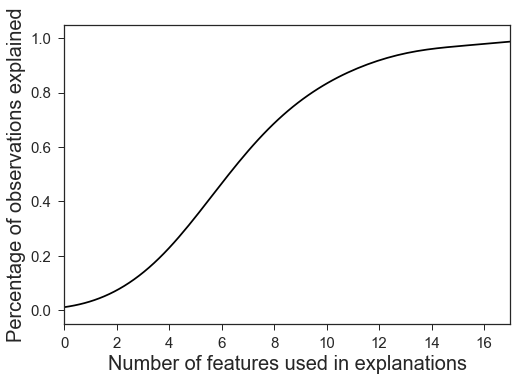} 
\par\end{centering}
\caption{Sparsity distribution over the news test dataset. Reading: '30\% of
the observations of our test dataset have explanations that use 5
features or less'.}

\label{dist_sparsity} 
\end{figure}

In order to check whether the proposed approach fulfills its goal
of finding explanations that can be easily understood by the user,
we evaluate the global sparsity of the explanations generated for
this problem. We measure sparsity as the number of non-zero coordinates
of the explanation vector $||x-e^{*}||_{0}$. Figure \ref{dist_sparsity}
shows the smoothed cumulative distribution of this value for all 11893
test data points. We observe that the maximum value over the whole
test dataset is 17, meaning that each observation of the test dataset
only needs to change 17 coordinates or less in order to cross the
decision boundary. Moreover, 80\% of them only need to move in 9 directions
or less, that is 15\% of the features only. This shows that the proposed
method indeed achieves sparsity in order to make explanations more
readable. It is important to note that this does not mean that we
only need 17 features to explain all the observations, since nothing
guarantees different explanations use the same features.

This experiment gives an illustration of how this method can be used
to gain knowledge on articles popularity prediction.

\subsection{Applications to Digit Classification}

Another application for this approach is to get some understanding
of how the model behaves in order to improve it. We use the MNIST
handwritten digits database \cite{LeCun1998} and apply \textit{Growing
Spheres} to the binary classification problem of recognizing the digits
8 and 9. The filtered dataset contains 11800 instances of 784 features
(28 by 28 pictures of digits). We use a support vector machine classifier
(\textbf{SVM}) with a RBF kernel and parameter $C=15$. We train the
model on 70\% of the data and test it on the rest (0.98 AUC score).
We use the same values for $\gamma$ and the hyperparameters of Algorithm
1 as in the first experiment.

\subsubsection{Illustrative Example}

\begin{figure}[t]
\begin{centering}
\includegraphics[scale=0.3]{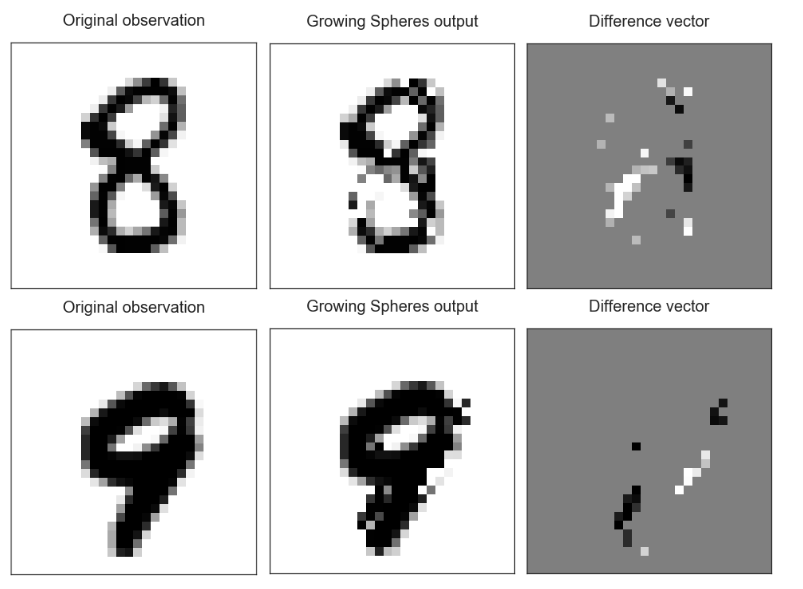}
\par\end{centering}
\caption{Output example from the application of \textit{Growing Spheres} for
two instances. Example of original instance $x$ (left column), its
closest ennemy found $e^{*}$(center) and the explanation vector $x-e^{*}$
(right). A white pixel indicates a 0 value, black a 1\label{fig:digit-example}}
\end{figure}

Given a picture of an 8 (Figure \ref{fig:digit-example}), our goal
is to understand how, according to the classifier, we could transform
this 8 into a 9 (and reciprocally), in order to get a sense of what
parts of the image are considered important. Our intuition would be
that 'closing the bottom loop' of a 9 should be the most influential
change needed to make a 9 become an 8, and hence features provoking
a class change should include pixels found in the bottom-left area
of the digits. Output examples to interprete a 9 and a 8 predictions
are shown in Figure \ref{fig:digit-example}.

Looking at Figure \ref{fig:digit-example}, the first thing we observe
confirms our intuition that a good proportion of the non-null coordinates
of the explanation vector are pixels located in the bottom-left part
of the digits (as seen in pictures right-column pictures). Hence,
we can see when comparing left and center pictures that \textit{Growing
Spheres} found the closest ennemies of the original observation by
either opening (top example) or closing (bottom example) the bottom
part of the digits. 

However, we also note that some pixels of the explanation vectors
are much harder to understand, such as the ones located on the top
right corner of the explanation image for instance. This was to be
expected since, as mentioned earlier, our method is trying to understand
the classifier's decision, not the reality it is approaximating. In
this case, the fact that the classifier apparently considers these
pixels to be influential the classification of these digits could
be an evidence of the learned boundary inaccuracy.

Finally, we note that the closest ennemies found by \textit{Growing
Spheres} (pictures in the center) in both cases are not proper 8 and
9 digits. Especially in the bottom example, a human observer would
still probably identify the center digit as a noised version of the
original 9 instead of an 8. Thus, despite achieving high accuracy
and having learned that bottom-left pixels are important to turn a
9 into an 8 and reciprocally, the classifier still fails to understand
the actual concepts making digits recognizable to a human. 

We also check the sparsity of our approach over the whole test set
(3528 instances). Once again, our method seems to be generating sparse
explanations since 100\% of the test dataset predictions can be interpreted
with explanations of at most 62 features (representing 7.9\% of total
features).

\section{Conclusion and Future Works}

The proposed post-hoc interpretability approach provides explanations
of a single prediction through the comparison of its associated observation
with its closest ennemy. In particular, we introduced a cost function
taking into account the sparsity of the explanations, and described
the implementation \textit{Growing Spheres,} which answers this problem
when having no information about the classifier nor existing data.
We showed that this approach provides insights about the classifier
through two applications. In the first one, \textit{Growing Spheres}
allowed us to gain meaningful information about features that were
locally relevant in news popularity prediction. The second application
highlighted both strengths and weaknesses of the support vector machine
used for digits classification, illustrating what concepts were learned
by the classifier. Furthermore, we also checked that the explanations
provided by the proposed approach are indeed sparse.

Beside collaborating with experts of industrial domains for explanations
validation, outlooks for our work include focusing on the constraints
imposed to the \textit{Growing Spheres} algorithm. In numerous real-world
applications, the final goal of the user may be such that it would
be useless for him to have explanations using specific features. For
instance, a business analyst using a model predicting whether or not
a specific customer is going to make a purchase would ideally have
an explanation based on features that he can leverage. In this context,
forbidding the algorithm to generate explanations in specific areas
of the input space or using specific features is a promising direction
for future work.

\bibliographystyle{aaai}
\bibliography{article_aies_arxiv}

\end{document}